\documentclass[letterpaper, 10 pt, journal, twoside]{ieeetran}
\usepackage{amsfonts}
\usepackage{algorithmic}
\usepackage{algorithm}
\usepackage{array}
\usepackage[caption=false,font=footnotesize]{subfig}

\usepackage{textcomp}
\usepackage{stfloats}
\usepackage{url}
\usepackage{verbatim}
\usepackage{graphicx}
\usepackage{cite}
\usepackage{booktabs}      % 三线表
\usepackage{siunitx}       % 数字对齐与单位处理
\usepackage{adjustbox}     % 使表格自适应宽度

\usepackage{amsmath}
\usepackage{microtype}
\allowdisplaybreaks

\usepackage{caption}
\usepackage{xcolor}

\definecolor{purple}{RGB}{128,0,128}

\graphicspath{{figures/}}

\usepackage{etoolbox} % 用于修改宏包内部命令

\graphicspath{{figures/}}
\usepackage[colorlinks=true]{hyperref}
\hypersetup{
    colorlinks=false,
	urlcolor=blue,          % URL链接的颜色
	citecolor=blue          % 引用链接的颜色
}

\begin{document}

\title{DRIFT: Diffusion-based Rule-Inferred For Trajectories}

\author{Jinyang Zhao, Handong Zheng, Yanjiu Zhong, Qiang Zhang, Shunyu Wu, Yu Kang

\thanks{
This work is supported by the National Natural Science Foundation of China (Number 72501093, 92367206, 92467302), 
% and the China Postdoctoral Science Foundation Anhui Joint Support Program under Grant Number 2024T012AH, and the China Postdoctoral Science Foundation under Grant Number 2025M781632
and the China Postdoctoral Science Foundation (2024T012AH, 2025M781632)
(Corresponding author:  Handong Zheng, Yanjiu Zhong.)}

\thanks{Jinyang Zhao, Handong Zheng, Yanjiu Zhong, Qiang Zhang and Yu Kang are with the School of Management, Ministry of Education Engineering Research Center for Intelligent Decision-Making \& Information System Technologies, Hefei University of Technology, Hefei, 230009, Anhui, China. Shunyu Wu is the Department of Automation, Shanghai Jiao Tong University, Shanghai 200240, China (e-mail: 2023213574@mail.hfut.edu.cn; zhenghandong@hfut.edu.cn; zhongyanjiu@sjtu.edu.cn;  qiang\_zhang@hfut.edu.cn; shunyuwu@sjtu.edu.cn; kangduyu@hfut.edu.cn).}
}


\markboth{Journal of \LaTeX\ Class Files,2026}%
{Shell \MakeLowercase{\textit{et al.}}: A Sample Article Using IEEEtran.cls for IEEE Journals}


\maketitle
\begin{abstract}

Trajectory generation for mobile robots in unstructured environments faces a critical dilemma: balancing kinematic smoothness for safe execution with terminal precision for fine-grained tasks. Existing generative planners often struggle with this trade-off, yielding either smooth but imprecise paths or geometrically accurate but erratic motions. To address the aforementioned shortcomings, this article proposes DRIFT (Diffusion-based Rule-Inferred for Trajectories), a conditional diffusion framework designed to generate high-fidelity reference trajectories by integrating two complementary inductive biases. First, a Relational Inductive Bias, realized via a GNN-based Structured Scene Perception (SSP) module, encodes global topological constraints to ensure holistic smoothness. Second, a Temporal Attention Bias, implemented through a novel Graph-Conditioned Time-Aware GRU (GTGRU), dynamically attends to sparse obstacles and targets for precise local maneuvering. In the end, quantitative results demonstrate that DRIFT reconciles these conflicting objectives, achieving centimeter-level imitation fidelity (0.041m FDE) and competitive smoothness (27.19 Jerk). This balance yields highly executable reference plans for downstream control.

\end{abstract}


\section{Introduction}

Generating safe, smooth, and feasible trajectories in large-scale, unstructured environments is a prerequisite for autonomous mobile robots~\cite{lavalle2006planning, thrun2002probabilistic}.
% While perception-driven mapless methods have rapidly advanced in generating feasible paths for open spaces~\cite{liang2024mtg, 10776999, 8460519, 7989381, mirowski2017learningnavigatecomplexenvironments}, the requirements for fine-grained tasks—such as docking or close-quarters maneuvering—remain challenging. 
While perception-driven mapless methods have rapidly advanced in producing feasible paths for open-space navigation~\cite{liang2024mtg, 10776999, 8460519, 7989381, mirowski2017learningnavigatecomplexenvironments}, fine-grained tasks such as docking, pre-grasping positioning, or close-quarters maneuvering remain particularly challenging.
In these scenarios, the validity of a trajectory is no longer defined merely by reachability, but by its ability to simultaneously satisfy two strict and often conflicting objectives: kinematic smoothness for safe execution and terminal precision for accurate task completion~\cite{mavrogiannis2023core, kruse2013human, 10305270}.

% The first requirement is Terminal Precision. For fine-grained tasks such as docking or pre-grasping positioning, the generated reference trajectory must satisfy strict endpoint constraints with high geometric accuracy. The second requirement is Kinematic Smoothness. To facilitate stable tracking by downstream low-level controllers and ensure predictability in shared spaces, the trajectory must minimize higher-order motion derivatives \cite{gao2022evaluation}. 
For fine-grained manipulation or docking tasks, the generated reference trajectory must converge to the target pose with high geometric precision under strict endpoint constraints. At the same time, to enable stable tracking by downstream low-level controllers and ensure predictability in shared environments, the trajectory must remain kinematically smooth by minimizing higher-order motion derivatives~\cite{gao2022evaluation}
However, these two objectives are often intrinsically conflicting in learning-based planning. 
% : enforcing strict terminal constraints typically induces high-frequency, erratic adjustments near the goal, thereby compromising smoothness.
Enforcing strict terminal constraints typically induces high-frequency adjustments near the goal, leading to erratic velocity or jerk profiles.
Conversely, prioritizing global smoothness often results in non-negligible terminal displacement errors. This precision-smoothness trade-off represents an unresolved fundamental bottleneck in current mapless trajectory generation approaches.

% Existing data-driven approaches typically bifurcate into two distinct paradigms regarding this trade-off. 
Existing data-driven methods generally address this trade-off through two distinct paradigms.
Classical generative models or optimization-embedded networks prioritize strict geometric constraints and terminal accuracy~\cite{4082128,  doi:10.1177/0278364911406761, 11106766}. 
% However, enforcing such rigid boundary conditions often induces kinematic discontinuities, yielding erratic profiles that are challenging for low-level controllers to track smoothly. 
However, effective in enforcing endpoint conditions, such rigid boundary constraints often introduce kinematic discontinuities, yielding trajectories that are difficult for low-level controllers to track smoothly.
In contrast, recent diffusion-based policies excel in synthesizing globally coherent and kinematically feasible motions, making them well-suited for unconstrained navigation~\cite{janner2022planningdiffusionflexiblebehavior,  doi:10.1177/02783649241273668,  sridhar2024nomad,  10802055, NEURIPS2020_4c5bcfec}. 
% Yet, these stochastic models inherently struggle with fine-grained conditioning, often failing to converge to the precise terminal states required for high-fidelity maneuvering. 
Nevertheless, due to their stochastic nature and reliance on static conditioning, these models commonly struggle with fine-grained terminal control, often failing to converge to the precise target states required for high-fidelity maneuvering.
Consequently, achieving "High-Fidelity Generation"—defined by the simultaneous satisfaction of kinematic smoothness and geometric precision—remains an open challenge.

% To bridge this gap,
To address this challenge, we propose DRIFT (Diffusion-based Rule-Inferred for Trajectories), a conditional generative framework for high-fidelity trajectory generation. 
% By embedding two complementary Architectural Inductive Biases directly into the diffusion process, DRIFT structurally decouples global topological reasoning from fine-grained local maneuvering, thereby simultaneously optimizing kinematic smoothness and terminal precision.
DRIFT explicitly incorporates architectural inductive biases into the diffusion process, structurally decoupling global topological reasoning from fine-grained local maneuvering. This design enables the planner to reconcile kinematic smoothness and terminal precision within a unified generative framework.

% The framework, besides, incorporates a Relational Inductive Bias via a GNN-based Structured Scene Perception (SSP) module to improve the global smoothness. 
Specifically, DRIFT integrates a Relational Inductive Bias through a GNN-based Structured Scene Perception (SSP) module to enhance global kinematic coherence.
Addressing the kinematic instability often caused by a lack of topological context, SSP encodes unstructured point clouds~\cite{NIPS2017_d8bf84be,  10.1145/3326362,  Hua_2018_CVPR} into a latent scene graph~\cite{rosinol20203d, 10933547,  10908637,  pmlr-v80-sanchez-gonzalez18a}. This structured representation explicitly captures environmental affordances, ensuring trajectories remain globally coherent and align proactively with the target pose, thereby structurally precluding high-Jerk artifacts.

Complementarily, to ensure terminal state fidelity, DRIFT introduces a Temporal Attention Bias via a novel Graph-Conditioned Time-Aware GRU (GTGRU). We attribute the terminal deviation in standard diffusion planners to their reliance on static conditioning, which fails to adapt to fine-grained constraints during the denoising process. By integrating Sparse Cross-Attention, GTGRU enables the generator to dynamically attend to critical obstacles and goal states at each timestep. This mechanism ensures that the generated reference trajectories converge precisely to the target pose while respecting local geometric constraints, effectively balancing precision and smoothness in cluttered environments~\cite{7989381, 9250507,  chen2019crowd, 10007923, 9197434, NIPS2017_3f5ee243}.

In summary, the main contributions of this work are as follows:

\begin{itemize} 
\item The proposal of DRIFT, a conditional diffusion-based trajectory generative framework that effectively reconciles the critical trade-off between geometric precision and kinematic smoothness for trajectory generation in unstructured environments.

\item The integration of a Relational Inductive Bias via a GNN-based structured perception module, which encodes global scene topology to ensure holistic kinematic coherence and traversability.

\item The design of a Temporal Attention Bias using a Graph-Conditioned Time-Aware GRU (GTGRU), which enables dynamic attention to fine-grained constraints for centimeter-level terminal fidelity. 

\end{itemize}

\section{Prior Work and Background}

\subsection{Structured Scene Encoding for Trajectory Generation}
The representation of the unstructured environment is fundamental to the fidelity of subsequent trajectory generation. Mainstream perception methods, such as semantic segmentation using CNNs or geometric reconstruction via Neural Radiance Fields (NeRFs), have achieved high fidelity in dense scene description~\cite{9712211, 10.1145/3503250, Cordts_2016_CVPR, Chen_2018_ECCV}. However, the architectural inductive biases of these grid-based or implicit representations—specifically their locality and translational invariance—are inherently ill-suited for modeling global topological relationships. Consequently, they struggle to capture long-range connectivity and spatial affordances required for consistent planning. To bridge this gap, GNNs offer a powerful paradigm. By explicitly modeling environmental entities as nodes and spatial constraints as edges, the unique relational inductive bias of GNNs enables the extraction of topologically consistent context embeddings from raw perception~\cite{battaglia2018relationalinductivebiasesdeep, rosinol20203d, 10933547, 10908637, pmlr-v202-liu23ax} providing a structured foundation for the downstream diffusion generator.

\subsection{Generative Models for Motion Planning} 

Learning-based trajectory generation has evolved from early Conditional Variational Autoencoders (CVAEs)~\cite{NIPS2015_8d55a249}, GANs~\cite{goodfellow2020generative,  Gupta_2018_CVPR, 10433735, 10138604} and Transformers~\cite{pmlr-v157-chen21a} to recent Denoising Diffusion Probabilistic Models (DDPMs)~\cite{doi:10.1177/02783649241273668, NEURIPS2020_4c5bcfec, liang2024mtg}. While early generative models and sequence predictors capture motion distributions, they often suffer from training instability or kinematic discontinuities.In contrast, diffusion-based planners, such as Diffuser~\cite{janner2022planningdiffusionflexiblebehavior} and DTG~\cite{10802055}, excel at synthesizing long-horizon, coherent trajectories. However, a critical bottleneck in existing diffusion policies is their reliance on static conditioning, where environmental context is encoded into a fixed vector prior to generation. This paradigm is generally ill-suited for fine-grained terminal tasks,
as it prevents the model from dynamically attending to local geometric constraints during the iterative denoising process.

\subsection{Optimizing Smoothness and Precision in Motion Planning}
Generating trajectories that effectively reconcile kinematic feasibility with geometric precision remains a fundamental challenge in unstructured environments. Traditional optimization-based planners (e.g., MPC)~\cite{4082128, doi:10.1177/0278364911406761, 11106766} excel at strictly enforcing terminal constraints and safety bounds. However, these approaches are often constrained by high computational latency and susceptibility to local minima. Conversely, learning-based policies (e.g., Behavior Cloning)~\cite{10305270, NEURIPS2024_da84e39a, doi:10.1177/0278364915619772} offer high inference efficiency but frequently struggle to capture the multi-modal nature of expert maneuvers. This limitation typically manifests as a trade-off between under-fitting—resulting in terminal spatial drift—and forced over-fitting, which induces high-frequency jitter that compromises downstream tracking stability. Currently, few frameworks effectively reconcile the conflict between kinematic coherence and terminal precision solely through the architectural inductive biases of a generative model, without reliance on computationally expensive online optimization. 
% This gap highlights the need for generative planning frameworks that can reconcile kinematic coherence and terminal precision without reliance on computationally expensive online optimization.

% In summary, prior work has made significant progress in structured scene representation, generative trajectory modeling, and motion optimization.
% While structured encodings offer rich topological context and diffusion-based planners generate globally coherent motions, their integration remains limited, particularly for enforcing fine-grained terminal constraints.
% This gap motivates a unified generative framework that combines structured scene representations with time-adaptive conditioning to reconcile kinematic smoothness and terminal precision.

In summary, prior work has made significant progress in structured scene representation, generative trajectory modeling, and motion optimization.
However, the lack of unified frameworks that explicitly integrate relational scene structure with time-adaptive generative processes continues to limit high-fidelity trajectory generation in fine-grained tasks.
This observation motivates the development of the approach introduced in the following section.

\section{PROBLEM FORMULATION}
\label{sec:problem_formulation}

\begin{figure*}[htbp]
  \centering
  \includegraphics[width=0.9\textwidth]{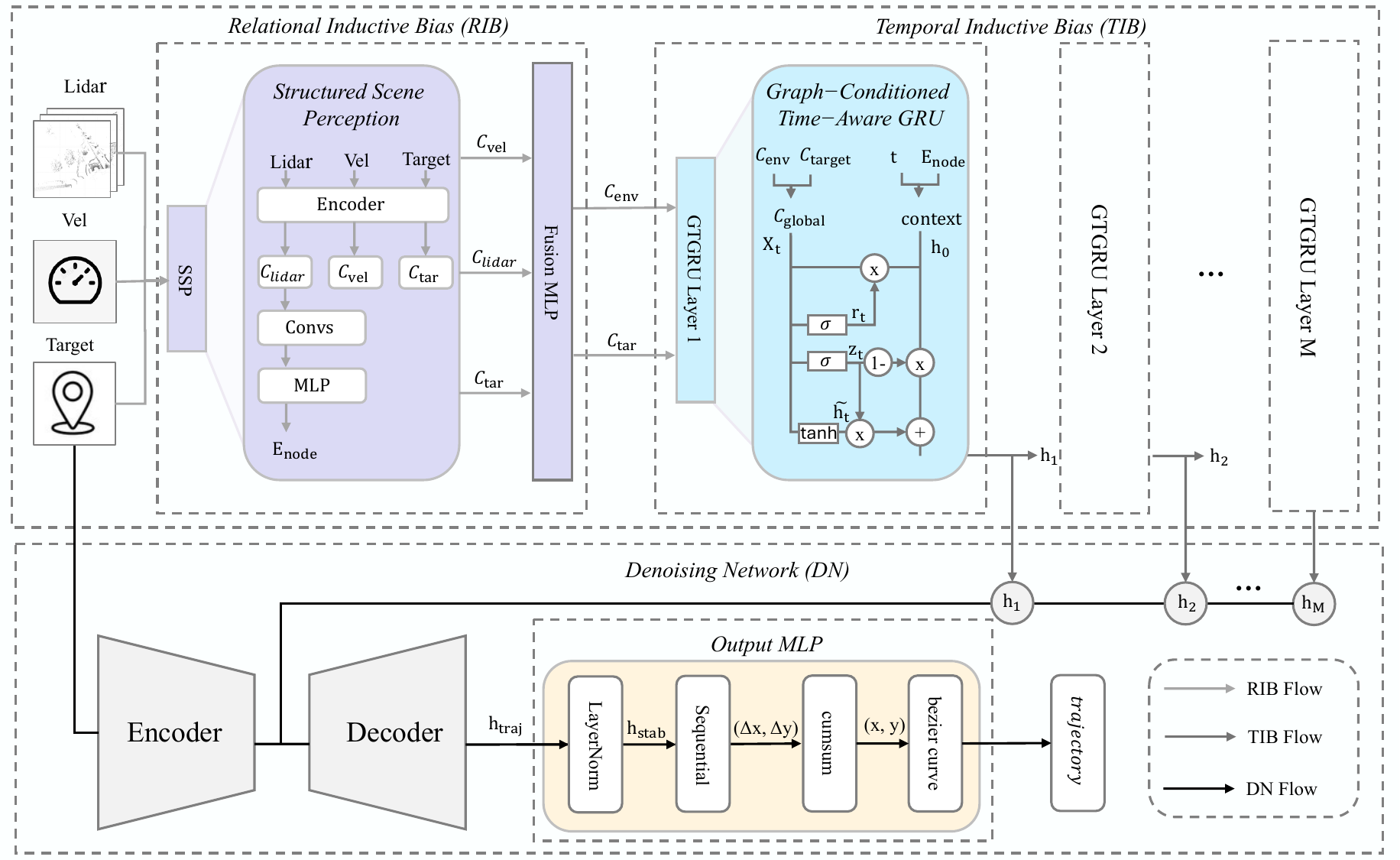}
  % --- 图注 ---
  \caption{DRIFT fuses relational topological bias (SSP) and temporal attention bias (GTGRU) to produce high-fidelity reference plans. The Denoising Network (DN) then iteratively refines trajectories from these structured encodings, balancing smoothness and precision.}
  \label{figure 1}
\end{figure*}

We formalize the mapless trajectory generation task as learning a conditional distribution $p_\theta(\tau | \mathcal{O})$ to synthesize high-fidelity reference plans. As illustrated in Fig.~\ref{figure 1}, at time step $t$, the observation $\mathcal{O}_t$ consists of the local LiDAR point cloud $\mathbf{P}_t \in \mathbb{R}^{N \times 3}$, the robot's historical velocity sequence $\mathbf{v}_t \in \mathbb{R}^{H_v \times 2}$, and the local target coordinates $\mathbf{g}_t \in \mathbb{R}^2$. The objective is to generate a future trajectory $\tau = (\mathbf{w}_1, \mathbf{w}_2, \ldots, \mathbf{w}_M)$, where $\mathbf{w}_i \in \mathbb{R}^2$ denotes the planned waypoint in the robot's local frame.

Unlike standard imitation learning, which solely minimizes geometric reconstruction errors (e.g., Hausdorff distance), a valid reference plan for fine-grained maneuvering must satisfy two sets of intrinsic constraints:
\begin{itemize}
    \item Global Topological Consistency ($\mathcal{C}_{global}$): The trajectory must be kinematically coherent and collision-free regarding the environmental topology. This corresponds to minimizing smoothness cost $\mathcal{L}_{smooth}$ and collision cost $\mathcal{L}_{coll}$.
    \item Terminal State Fidelity ($\mathcal{C}_{local}$): The trajectory must precisely converge to the target state with minimal displacement, necessitating the minimization of terminal error $\mathcal{L}_{term}$.
\end{itemize}

Mathematically, let $\hat{\tau} = \pi_\theta(\mathcal{O}_t)$ denote the trajectory generated by the policy. The optimal parameters $\theta^*$ are learned by minimizing a compound objective function:
\begin{equation}
\label{eq:objective}
\begin{aligned}
    \theta^* = \arg\min_{\theta} \mathbb{E}_{\tau_{gt} \sim \mathcal{D}} \Big[ &\mathcal{L}_{\text{imit}}(\hat{\tau}, \tau_{gt}) \\
    &+ \lambda_1 \mathcal{C}_{global}(\hat{\tau}) + \lambda_2 \mathcal{C}_{local}(\hat{\tau}) \Big],
\end{aligned}
\end{equation}
where $\lambda_{1,2}$ are dynamic weighting factors balancing the constraints.

However, directly optimizing Eq.~\eqref{eq:objective} is challenging due to the non-differentiable nature of obstacle avoidance and the conflict between smoothness and precision. To address this, DRIFT adopts an ``Architecture-as-Regularizer'' approach. Instead of relying solely on loss constraints, we embed specific inductive biases directly into the network architecture. Specifically, the SSP module injects a Relational Inductive Bias to implicitly satisfy $\mathcal{C}_{global}$ by encoding scene topology $E_{node}$, while the Graph-Conditioned Time-Aware GRU (GTGRU) injects a Temporal Attention Bias to enforce $\mathcal{C}_{local}$ by dynamically attending to fine-grained terminal constraints.

\section{METHODOLOGY}

\subsection{Overview}
The proposed DRIFT framework reformulates mapless navigation planning as a conditional trajectory generation process, explicitly decoupling the challenge into global topological reasoning and local fine-grained refinement. To reconcile the inherent trade-off between kinematic smoothness and terminal precision, the architecture integrates two orthogonal Architectural Inductive Biases. First, a SSP module injects a Relational Inductive Bias, encoding unstructured point clouds into a latent scene graph to enforce global topological consistency. Second, a Recurrent Denoising Generator (RDG) incorporates a Temporal Attention Bias via an autoregressive denoising mechanism. This generator is instantiated by a novel Graph-Conditioned Time-Aware GRU (GTGRU), enabling the model to dynamically attend to critical terminal constraints for high-fidelity convergence.

\subsection{Relational Inductive Bias: Structured Scene Perception}

To instantiate the Relational Inductive Bias, the SSP module is designed to extract topologically consistent representations from unstructured inputs. Unlike projection-based methods that compress spatial data into global vectors, SSP explicitly models the local environment as a latent graph $\mathcal{G} = (\mathcal{V}, \mathcal{E})$ to reason about geometric affordances.

The pipeline begins by processing the raw LiDAR point cloud $\mathbf{P}_t$. To balance computational efficiency with local detail, we apply voxel-based downsampling to discretize the cloud into local regions, treating each populated voxel as a graph node $v_i \in \mathcal{V}$ (Fig. \ref{fig:sparsification}). The initial node feature $\mathbf{x}_i$ encodes the centroid coordinates and local point density. Topological connectivity is established by constructing edges $\mathcal{E}$ via $k$-Nearest Neighbors ($k$-NN) in the feature space, ensuring robust message passing even in sparse regions.

This latent graph is processed by a GNN backbone composed of stacked EdgeConv layers~\cite{10.1145/3326362}. It is selected for its ability to capture local geometric structures by operating on edge features. The node feature update is defined as an symmetric aggregation over local neighborhoods $\mathcal{N}(v_i)$:
\begin{equation}
    \mathbf{h}_i^{(l+1)} = \max_{j \in \mathcal{N}(v_i)} \text{MLP}^{(l)} \left( \mathbf{h}_i^{(l)} \oplus (\mathbf{h}_j^{(l)} - \mathbf{h}_i^{(l)}) \right),
\end{equation}
where $\oplus$ denotes feature concatenation, and $\mathbf{h}_i^{(l)}$ represents the hidden feature vector of node $v_i$ at the $l$-th layer (with $\mathbf{h}_i^{(0)} = \mathbf{x}_i$). The term $(\mathbf{h}_j^{(l)} - \mathbf{h}_i^{(l)})$ explicitly captures the relative spatial relationship between the centroid node $i$ and its neighbor $j$. This operation ensures that the learned representations capture local geometric details invariant to global translation.

Critically, the SSP module yields two distinct representations to support the dual constraints of our framework:
\begin{itemize}
    \item Node-level Embeddings ($E_{node}$): The dense features from the final GNN layer preserve fine-grained spatial details, which are retained for the subsequent attention mechanism to query local obstacles.

\begin{algorithm}[H]
\caption{Structured Scene Perception (SSP) Module}\label{alg:ssp_module}
\begin{algorithmic}[1]
\STATE \textsc{GenerateGraphEmbeddings}$(\mathcal{O}_l; k, L)$
\STATE \textbf{Inputs:} raw point set / observations $\mathcal{O}_l$, neighbor size $k$, GNN depth $L$
\STATE \textbf{Outputs:} node embeddings $E_{\text{node}}$, global context $c_{\text{lidar}}$
\STATE \textbf{1. Dynamic Graph Construction}
\STATE $V_{\text{sparse}} \leftarrow \mathrm{VoxelDownsample}(\mathcal{O}_l)$
\STATE $\mathcal{E} \leftarrow \mathrm{k\text{-}NN}(V_{\text{sparse}}, k)$
\STATE $h^{(0)} \leftarrow \mathrm{InitialFeatures}(V_{\text{sparse}})$
\STATE $\mathcal{G} \leftarrow (V_{\text{sparse}}, h^{(0)}, \mathcal{E})$ 
\STATE \textbf{2. GNN Feature Propagation (RGE)}
\STATE \textbf{for} $l = 0$ \textbf{to} $L-1$ \textbf{do}
\STATE \textbf{for} each node $v \in V_{\text{sparse}}$ \textbf{do}
\STATE $h_{\text{edge}} \leftarrow \bigoplus_{u \in \mathcal{N}(v)} \mathrm{MLP}_{\phi}^{(l)}\big(h_v^{(l)} \oplus (h_u^{(l)} - h_v^{(l)})\big)$
\STATE $h_v^{(l+1)} \leftarrow f_{\theta}^{(l)}\big(h_v^{(l)},\, h_{\text{edge}}\big)$
\STATE \textbf{end for}
\STATE \textbf{end for}
\STATE \textbf{3. Generate Outputs}
\STATE $E_{\text{node}} \leftarrow \{h_v^{(L)}\}_{v\in V_{\text{sparse}}}$ \quad 
\STATE $c_{\text{lidar}} \leftarrow \mathrm{GlobalMaxPool}(E_{\text{node}})$ \quad 
\STATE \textbf{return} $E_{\text{node}},\; c_{\text{lidar}}$
\end{algorithmic}
\end{algorithm}
    
    \item Environmental Context ($c_{env}$): A global context vector is derived by fusing the pooled graph feature $c_{lidar} = \text{MaxPool}(E_{node})$ with the robot's historical kinematic state. The velocity history $\mathbf{v}_t$ is encoded into $c_{vel}$ via a 1D-CNN, and fused as:
    \begin{equation}
    \label{eq:c_env}
        c_{env} = \text{MLP}_{\text{fuse}}(c_{lidar} \oplus c_{vel}).
    \end{equation}
\end{itemize}
Notably, unlike prior works that bake the goal into the global context, we deliberately exclude the target $g_t$ from $c_{env}$. This design enforces a structural decoupling, ensuring that the perception module focuses purely on environmental traversability (satisfying $\mathcal{C}_{global}$), while target guidance is handled dynamically by the temporal generation process.

\subsection{Temporal Inductive Bias: Recurrent Denoising Generator}
\label{subsec:rdg}

Complementarily, to enforce Terminal State Fidelity ($\mathcal{C}_{local}$), the framework introduces a Temporal Attention Bias via the Recurrent Denoising Generator (RDG). While the SSP module provides a static global map, the diffusion denoising process requires dynamic adaptability to fine-grained constraints that vary along the prediction horizon. The RDG employs a novel Graph-Conditioned Time-Aware GRU (GTGRU) as its backbone, reformulating trajectory generation as a sequence-to-sequence translation task conditioned on topological embeddings.

Specifically, at each diffusion step $k$, the noisy trajectory $\mathbf{x}^{(k)}$ is processed autoregressively. For the $m$-th waypoint $\mathbf{w}_m$, the GTGRU cell updates its hidden state $\mathbf{h}_m$ by integrating three critical information streams:

\begin{figure}[!t]
  \centering
  \includegraphics[height=130pt, width =0.7\columnwidth]{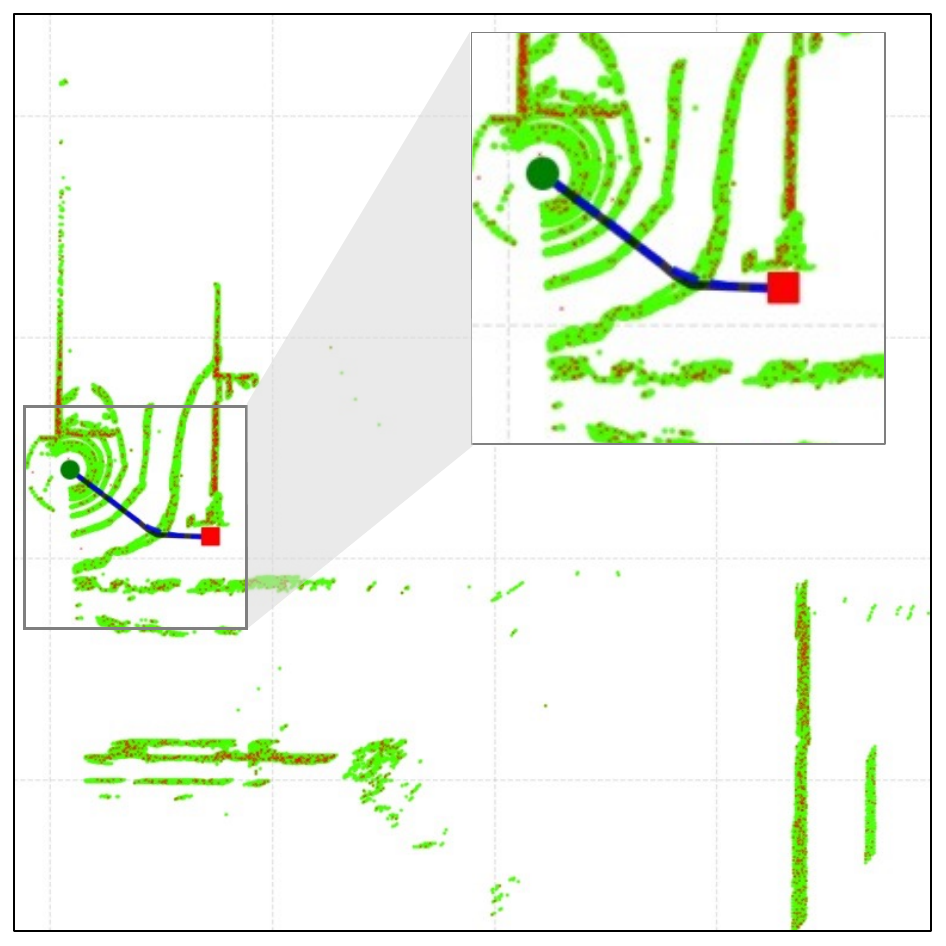} % 改为 0.8、0.6 等
    \caption{Voxel downsampling converts the dense LiDAR point cloud (\textcolor{green}{green}) into sparse graph nodes (\textcolor{red}{red}) (Alg. \ref{alg:ssp_module}), cutting computational cost and producing structured topologies for GNN relational reasoning.}
\label{fig:sparsification}
\end{figure}

\subsubsection{Dynamic Obstacle Attention}
To resolve local collision risks, we employ a Sparse Cross-Attention mechanism. Unlike static pooling, the current hidden state $\mathbf{h}_{m-1}$ serves as a query to retrieve relevant spatial features from the node embeddings $E_{node}$ (Key/Value) generated by SSP. This produces a dynamic context vector $\mathbf{d}_m$ that focuses specifically on obstacles relevant to the current horizon step:
\begin{equation}
    \mathbf{d}_m = \text{Attention}(\mathbf{Q}=\mathbf{h}_{m-1}, \mathbf{K}=E_{node}, \mathbf{V}=E_{node}).
\end{equation}

This mechanism allows the planner to "look at" specific graph nodes (e.g., a door frame or obstacle edge) only when the trajectory approaches them, ensuring kinematic clearance without over-smoothing.

\subsubsection{Target-Driven Gating}
To ensure precise convergence to the goal, the target encoding $c_{target}$ is fused with the global context $c_{env}$ and time embedding $t_{embed}$ to form a conditioning vector $\mathbf{c}_{cond}$. This vector is injected into every recurrent step, directly modulating the GRU's update gate $\mathbf{z}_m$ and reset gate $\mathbf{r}_m$:
\begin{equation}
    \begin{aligned}
        \mathbf{z}_m &= \sigma(W_z \mathbf{w}_m + U_z \mathbf{h}_{m-1} + V_z \mathbf{d}_m + C_z \mathbf{c}_{cond}), \\
        \mathbf{h}_m &= (1 - \mathbf{z}_m) \odot \mathbf{h}_{m-1} + \mathbf{z}_m \odot \tilde{\mathbf{h}}_m.
    \end{aligned}
\end{equation}

By persistently injecting $c_{target}$ into the gating mechanism, the network maintains a strong gradient flow towards the terminal state throughout the long-horizon generation, effectively minimizing the endpoint displacement error $\mathcal{L}_{term}$.

\subsection{Training Objectives and Curriculum}
\label{subsec:training}

To enforce the proposed architectural inductive biases, we adopt a prediction-based training strategy. Unlike standard diffusion models that predict the noise residue $\epsilon$, our generator $G_\theta$ is parameterized to predict the noise-free trajectory $\hat{\tau} = \mathbf{x}_0$ directly. This formulation allows us to apply differentiable geometric and kinematic constraints directly in the output space.

The global objective $\mathcal{L}_{total}$ combines the standard diffusion denoising loss $\mathcal{L}_{simple}$ with a set of auxiliary constraint losses $\mathcal{L}_{aux}$, dynamically balanced by a curriculum scheduler:
\begin{equation}
    \mathcal{L}_{total} = \mathcal{L}_{simple}(\epsilon, \epsilon_\theta) + \sum_{k} \lambda_k(e) \cdot \mathcal{L}_k(\hat{\tau}, \tau_{gt}, \mathcal{O}_t),
\end{equation}
where $e$ denotes the current epoch. The auxiliary losses are designed to enforce the dual constraints defined in Sec.~\ref{sec:problem_formulation}:

\subsubsection{Imitation Fidelity ($\mathcal{L}_{imit}$)} 
To satisfy Terminal State Fidelity ($\mathcal{C}_{local}$) and capture expert intent, we minimize a compound geometric distance comprising Hausdorff distance, Dynamic Time Warping (DTW), and endpoint MSE:
\begin{equation}
    \mathcal{L}_{imit} = \mathcal{L}_{Haus}(\hat{\tau}, \tau_{gt}) + \mathcal{L}_{DTW}(\hat{\tau}, \tau_{gt}) + \|\hat{\mathbf{w}}_M - \mathbf{g}_t\|^2.
\end{equation}

\subsubsection{Kinematic Coherence ($\mathcal{L}_{smooth}$)} 
To ensure Global Topological Consistency ($\mathcal{C}_{global}$), we explicitly penalize high-frequency jitter. The smoothness loss maximizes the cosine similarity between adjacent velocity vectors $\mathbf{v}_i = \mathbf{w}_{i+1} - \mathbf{w}_i$ to encourage directional continuity:
\begin{equation}
    \mathcal{L}_{smooth} = \mathbb{E}_{i} \left[ 1 - \frac{\mathbf{v}_i \cdot \mathbf{v}_{i+1}}{\|\mathbf{v}_i\|_2 \|\mathbf{v}_{i+1}\|_2 + \xi} \right].
\end{equation}

\subsubsection{Safety Constraint ($\mathcal{L}_{coll}$)} 
We impose a soft penalty for any waypoint $\mathbf{w}_i$ that breaches the safety radius $r_{safe}$ of obstacle points $\mathbf{o}_j \in \mathbf{P}_t$:
\begin{equation}
    \mathcal{L}_{coll} = \sum_{i,j} \max\left(0, r_{safe} - \|\mathbf{w}_i - \mathbf{o}_j\|_2\right)^2.
\end{equation}

\subsubsection{Curriculum Strategy} 
We implement a two-stage curriculum via $\lambda_k(e)$. In the initial Imitation Phase ($e < E_{stage1}$), weights prioritize $\mathcal{L}_{imit}$ to establish a valid trajectory manifold. In the subsequent Refinement Phase ($e \ge E_{stage1}$), we significantly increase the weights for $\mathcal{L}_{smooth}$ and $\mathcal{L}_{coll}$ (e.g., increasing $\lambda_{coll}$ by $3\times$). This strategy guides the model from learning basic reachability to refining high-fidelity, collision-free plans suitable for downstream control.

\begin{figure}[!t] 
  \centering
  \includegraphics[width=\columnwidth]{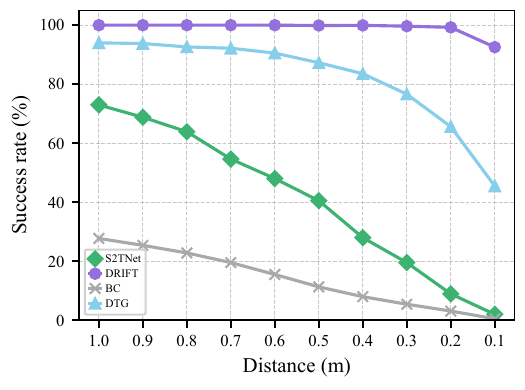}
  \caption{Terminal accuracy comparison for close-proximity docking. Our method, DRIFT (\textcolor{purple}{purple}), maintains 92.55\% success at 0.1m, while the DTG baseline (\textcolor{blue}{blue}) fails (45.49\%). This result demonstrates DRIFT's 
  ability to resolve the "precision-smoothness trade-off" by achieving high-precision local maneuvering.
  }
  \label{fig:wide}
\end{figure}

\section{EXPERIMENTS}
\label{sec:experiments}

\begin{figure*}[t] % 来确保它在页面顶部
  \centering
  \includegraphics[height=100pt, width =1.0\textwidth]{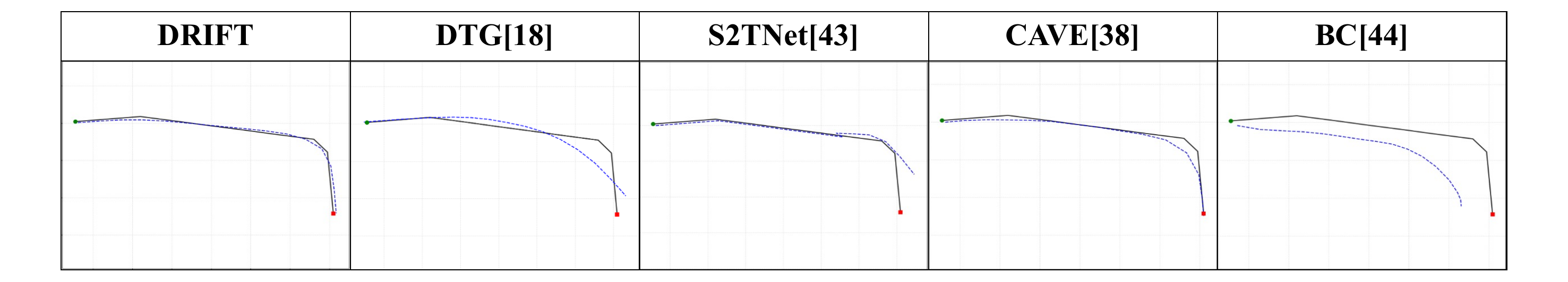}
  \caption{Predicted trajectories (dark) vs. ground truth (light gray) across models: DTG is smooth but deviates and misses terminals; CAVE and BC are erratic or imprecise. DRIFT resolves the precision–smoothness trade-off, matching the ground truth while remaining smooth.}
  \label{fig:wide}
\end{figure*}

\subsection{Experimental Setup}

We utilize the large-scale outdoor navigation dataset from \cite{liang2024mtg}, collected by a Husky robot equipped with a 3Hz VLP-16 LiDAR. Following the standard protocol \cite{liang2024mtg}, the observation input comprises the last $C_l=3$ LiDAR frames and $C_v=10$ velocity frames. Ground truth trajectories are generated via an A* planner, thresholded at 15m with $H=16$ waypoints. Goals are sampled within 60m for training and $>50$m for testing to evaluate long-horizon generation. Unlike the DTG baseline which uses PointCNN with 0.08m voxels, our perception encoder employs a coarser 0.2m voxel downsampling coupled with a $k$-NN ($k=8$) graph constructor to explicitly preserve topological connectivity. All models are trained on an NVIDIA RTX 3090 GPU.

\begin{algorithm}[H]
\caption{DRIFT Denoising Step with GTGRUCell}
\label{alg:gtgru_step_tanimoto}
\begin{algorithmic}[1]
\STATE \textsc{DenoisingStep}$(h_{\mathrm{prev}}, x_{\mathrm{prev}}, p_t^{\mathrm{curr}}, E_{\mathrm{node}}, c_{\mathrm{target}}, t_{\mathrm{embed}})$
\STATE \textbf{1. Sparse Cross-Attention}
\STATE $Q \leftarrow \mathrm{MLP}_Q(h_{\mathrm{prev}})$
\STATE $K, V \leftarrow \mathrm{MLP}_K(E_{\mathrm{node}}),\ \mathrm{MLP}_V(E_{\mathrm{node}})$
\STATE $\mathcal{N}_k \leftarrow \mathrm{k\text{-}NN}(p_t^{\mathrm{curr}}, E_{\mathrm{node}}, k_{\mathrm{env}})$
\STATE $\alpha \leftarrow \mathrm{Softmax}\big(\mathrm{Similarity}(Q, K[\mathcal{N}_k])\big)$
\STATE $d_i \leftarrow \sum_{j\in\mathcal{N}_k} \alpha_j V_j$
\STATE \textbf{2. Gated Recurrent Unit (GTGRU) Update}
\STATE $x_{\mathrm{in}} \leftarrow \mathrm{PointEmbed}(x_{\mathrm{prev}}) + \mathrm{TargetProj}(c_{\mathrm{target}})$
\STATE $z_t \leftarrow \sigma\big(W_z x_{\mathrm{in}} + U_z h_{\mathrm{prev}} + V_z d_i + \dots \big)$ 
\STATE $r_t \leftarrow \sigma\big(W_r x_{\mathrm{in}} + U_r h_{\mathrm{prev}} + V_r d_i + \dots \big)$
\STATE $\tilde{h}_i \leftarrow \tanh\big(W_h x_{\mathrm{in}} + U_h (r_t \odot h_{\mathrm{prev}}) + \dots \big)$
\STATE $h_i \leftarrow (1 - z_t) \odot h_{\mathrm{prev}} + z_t \odot \tilde{h}_i$
\STATE \textbf{3. Waypoint Prediction}
\STATE $w_i^0 \leftarrow \mathrm{MLP}_{\mathrm{out}}(h_i)$
\STATE \textbf{return} $h_i, w_i^0$
\end{algorithmic}
\end{algorithm}

\subsubsection{Dataset and Challenge Protocols}
The dataset is partitioned into spatially disjoint training and testing areas. To rigorously stress-test the generation fidelity, we constructed a \textbf{Challenge Subset} by filtering the test split based on four extreme geometric and kinematic thresholds:
\begin{itemize}
    \item Spatial Constriction: Traversable width $< 2.0$m, demanding precise boundary adherence.
    \item High Clutter: Obstacle occupancy $> 15\%$ ($3.0$m radius), requiring intricate path generation.
    \item High Dynamics: Expert trajectory Jerk $> 5.0 \text{m/s}^3$, representing aggressive evasion maneuvers.
    \item Topological Complexity: Expert-to-A* length ratio $\gg 1$, implying latent navigation rules beyond Euclidean heuristics.
\end{itemize}

\subsubsection{Baselines}
We compare DRIFT against representative methods from distinct learning paradigms to validate its superiority in the mapless trajectory generation task. Optimization-based planners (e.g., MPC) are excluded as they rely on explicit mapping, which falls outside the scope of this end-to-end learning study.
\begin{itemize}
    \item BC (Behavior Cloning)~\cite{NEURIPS2024_da84e39a}: A fundamental imitation learning baseline with a deterministic perception-to-action mapping. While recent theoretical analysis suggests its optimality under certain conditions, in our context, it serves as a standard benchmark for geometric fitting performance.
    \item CAVE~\cite{NIPS2015_8d55a249} (Conditional VAE): A classic generative baseline. In trajectory synthesis, VAEs typically produce geometrically diverse but kinematically incoherent (high-Jerk) paths, representing the ``high-precision, low-smoothness'' extreme.
    \item S2TNet~\cite{pmlr-v157-chen21a} (Spatio-Temporal Transformer)~\cite{pmlr-v157-chen21a}: A SOTA trajectory predictor utilizing transformers to model complex interactions. It serves as a high-performance baseline for precision-focused architectures.
    \item DTG~\cite{10802055} (Diffusion-based): A SOTA diffusion policy for navigation. While excelling at generating smooth, naturalistic paths, it relies on static conditioning, often lacking the fine-grained terminal fidelity required for docking tasks. It represents the high-smoothness, low-precision'' extreme.
\end{itemize}
\subsubsection{Evaluation Metrics}
To rigorously quantify the quality of the generated reference plans $\tau$, we employ a comprehensive set of metrics across three dimensions:

\begin{itemize}
    \item Geometric Fidelity: 
    \emph{Final Displacement Error (FDE)} measures the Euclidean distance between the last generated waypoint $\mathbf{w}_M$ and the local goal $g$, quantifying the terminal precision ($\mathcal{C}_{local}$). 
    \emph{Length Fidelity (LF)} computes the ratio between the generated trajectory length and the ground truth length, quantifying the model's ability to replicate the expert's geometric scale without taking shortcuts or generating redundant motions.

    \item Kinematic Coherence: 
    \emph{Average Jerk ($\mathcal{J}$)} evaluates the smoothness of the trajectory by calculating the mean magnitude of the third derivative of position. Lower values indicate smoother, more trackable motions, reflecting global topological consistency ($\mathcal{C}_{global}$).

    \item Generation Feasibility: 
    \emph{Inference Success Rate (ISR)} reports the percentage of episodes where the generator produces a collision-free path that converges to the goal region (error $< 0.5m$) within the horizon.
    \emph{Predicted Collision Rate (PCR)} measures the percentage of generated trajectories that intersect with obstacle points in the input point cloud $\mathcal{O}_l$. 
    \emph{Inference Latency} records the average wall-clock time for a single generation step.
\end{itemize}
% --- 代码结束 ---

\subsection{Quantitative Analysis}
The quantitative results (Table~\ref{tab:results}) empirically validate the proposed ``Architecture-as-Regularizer'' paradigm and reveal the limitations of existing baselines in reconciling the fidelity-smoothness trade-off.

\subsubsection{Analysis of the Fidelity-Smoothness Dilemma}
The baselines exhibit a stark polarization, confirming the conflicting nature of geometric and kinematic constraints:
\begin{itemize}
    \item High Fidelity, Low Smoothness (CAVE): While CAVE achieves exceptional geometric fidelity with a low FDE ($0.041$m), this precision comes at the cost of severe kinematic instability, evidenced by a catastrophic Jerk ($527.87$ m/s$^3$). Such erratic, high-frequency oscillations render the generated trajectories practically untrackable for downstream controllers.
    
    \item High Smoothness, Low Fidelity (DTG): Conversely, the diffusion-based DTG produces the smoothest trajectories ($22.49$ m/s$^3$) but suffers from significant terminal drift ($0.699$m FDE) and a high Predicted Collision Rate ($32.47\%$). This validates our hypothesis that static conditioning fails to provide the fine-grained guidance necessary for terminal state convergence.
    
    \item Suboptimal Baselines (S2TNet \& BC): Standard imitation methods struggle to master either objective, yielding trajectories with both high displacement errors ($>0.7$m) and poor kinematic coherence.
\end{itemize}

\subsubsection{Reconciling the Conflict with DRIFT}
DRIFT effectively resolves this dilemma by structurally decoupling global and local reasoning. It achieves centimeter-level terminal fidelity ($0.041$m FDE), comparable to the precision-focused CAVE, while simultaneously maintaining competitive kinematic smoothness ($27.19$ m/s$^3$), on par with the smoothness-focused DTG. Crucially, this balance leads to a high Inference Success Rate ($91.66\%$) and a low Predicted Collision Rate ($5.17\%$). These results demonstrate DRIFT's unique capability to generate reference plans that are both geometrically accurate (for fine-grained tasks) and kinematically coherent (for stable execution), fulfilling the core requirements of high-fidelity trajectory generation.
% --- 单栏图片 ---
\begin{figure}[!t]
\centering
\includegraphics[width=\columnwidth]{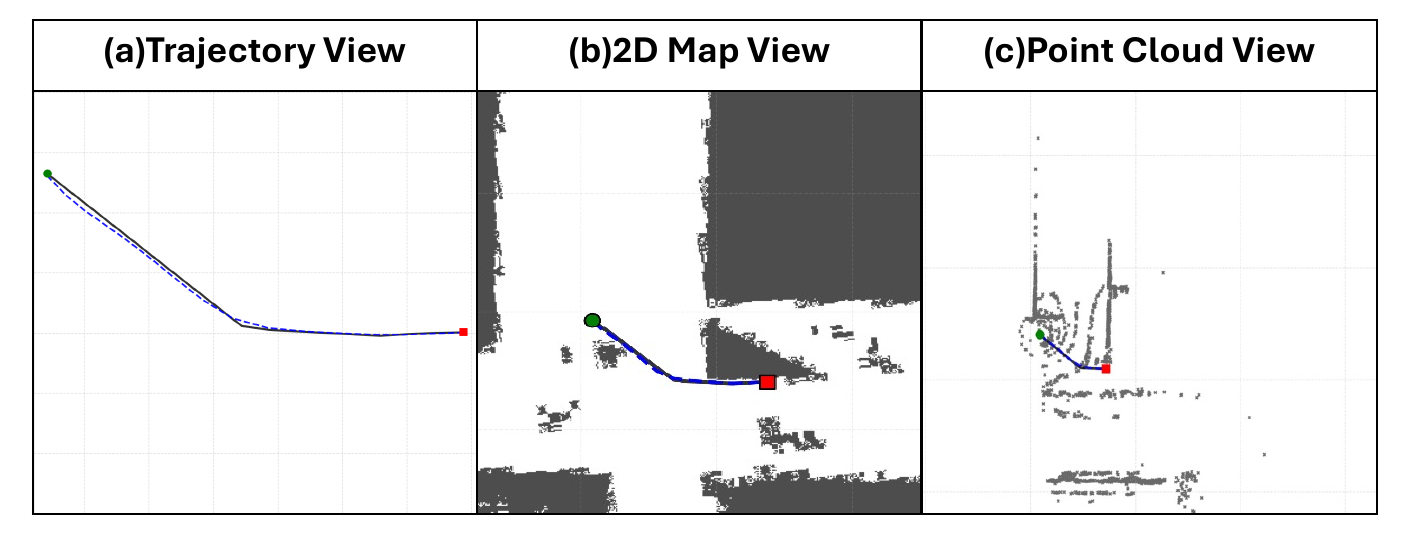}

% --- 图例 ---
\caption{Predicted trajectory (dark) closely matches ground truth (light gray) in (a) trajectory, (b) 2D map, and (c) point-cloud views, showing DRIFT reconciles precision and smoothness.}

\label{fig:traj_comparison}
\end{figure}
% --- 单栏图片代码结束 ---

\begin{table*}[!t]
    \centering
    \caption{QUANTITATIVE EVALUATION OF TRAJECTORY GENERATION PERFORMANCE}
    \label{tab:results}
    
    \begin{tabular*}{\textwidth}{l @{\extracolsep{\fill}} cccccc}
        \toprule
        % Updated Headers: More precise terminology
        Methods & ISR (\%)  & PCR (\%)  & Len. Fid. (\%)  & FDE (m)  & Jerk (m/s$^3$)  & Latency (s)  \\
        \midrule
        
        BC~\cite{NEURIPS2024_da84e39a}  & 0.52 & 5.55 & 90.61 & 1.44 & 118.37 & \textbf{0.05} \\
        CAVE~\cite{NIPS2015_8d55a249}  & \textbf{96.80} & \textbf{4.73} & 99.72 & \textbf{0.041} & 527.87 & 0.09 \\
        S2TNet~\cite{pmlr-v157-chen21a}  & 2.35 & 5.32 & \textbf{99.73} & 0.712 & 163.19 & 0.08 \\
        DTG~\cite{10802055}  & 45.35 & 32.47 & 99.20 & 0.699 & \textbf{22.49} & 0.14 \\
        \midrule
        \textbf{DRIFT (Ours)}  & 91.66 & 5.17 & 99.73 & \textbf{0.041} & 27.19 & 0.27 \\

        w/o GNN  & 87.45 & 5.40 & 99.71 & 0.058 & 22.17 & 0.30 \\
        w/o GTGRU (U-Net)  & 92.03  & 5.47 & 99.70 & 0.043 & 27.84 & 2.10 \\
        w/o Attention  & 64.04 & 5.51 & 99.82 & 0.094 & 24.05 & 0.29 \\
        \bottomrule
    \end{tabular*}
    
    \vspace{1ex}
    \raggedright
    \footnotesize{\textit{Note:  \textbf{ISR}: Inference Success Rate, \textbf{PCR}: Predicted Collision Rate, \textbf{Len. Fid.}: Length Fidelity (ratio to ground truth length). DRIFT achieves the best balance between geometric precision (FDE) and kinematic smoothness (Jerk).}}
\end{table*}
\subsection{Ablation Analysis: Inductive Biases}
To validate the contribution of the proposed architectural biases, the variants presented in Table~\ref{tab:results} are analyzed.

\subsubsection{Impact of Relational Inductive Bias (w/o GNN)}
Replacing the SSP module with a standard U-Net encoder results in a $4.2\%$ drop in ISR. This performance degradation suggests that grid-based convolutions struggle to capture the non-local topological connectivity of unstructured point clouds, which is critical for resolving global geometric constraints in complex environments.

\subsubsection{Impact of Temporal Attention Bias (w/o Attention)}
Removing the dynamic attention mechanism (GTGRU) leads to a catastrophic performance drop, with ISR falling to $64.04\%$ and FDE degrading significantly to $0.094$m. This empirical evidence validates the hypothesis that static conditioning is insufficient for fine-grained tasks. The per-step attention mechanism is indispensable for correcting accumulation errors during the denoising process, thereby ensuring terminal state fidelity.

\section{Conclusion}
In this letter, we presented DRIFT, a conditional generative framework that effectively reconciles the trade-off between geometric precision and kinematic smoothness for mapless trajectory generation. By adopting an ``Architecture-as-Regularizer'' paradigm, we embedded Relational and Temporal Inductive Biases to decouple global topological reasoning from fine-grained terminal refinement. Quantitative experiments demonstrate that DRIFT achieves centimeter-level imitation fidelity while maintaining competitive smoothness, surpassing existing baselines in generating executable reference plans. Future work will focus on accelerating inference via Consistency Models and integrating the generator with closed-loop tracking controllers for dynamic real-world deployment.
\section*{Acknowledgments}
This should be a simple paragraph before the References to thank those individuals and institutions who have supported your work on this article.
\bibliographystyle{IEEEtran}
\bibliography{Reference}

\vfill

\end{document}